# Automatic Identification of Closely-related Indian Languages : Resources and Experiments

Ritesh Kumar[1], Bornini Lahiri[2], Deepak Alok[3], Atul Kr. Ojha[4], Mayank Jain[4], Abdul Basit[1], Yogesh Dawer[1]

[1]Dr. Bhim Rao Ambedkar University, [2]Jadavpur University, [3]Rutgers University, [4]Jawaharlal Nehru University
Agra, Kolkata, USA, Delhi
{riteshkrjnu, lahiri.bornini, deepak06alok, shashwatup9k, jnu.mayank, basit.ansari.in, yogeshdawer}@gmail.com

**Abstract**

In this paper, we discuss an attempt to develop an automatic language identification system for 5 closely-related Indo-Aryan languages of India – Awadhi, Bhojpuri, Braj, Hindi and Magahi. We have compiled a comparable corpora of varying length for these languages from various resources. We discuss the method of creation of these corpora in detail. Using these corpora, a language identification system was developed, which currently gives state-of-the-art accuracy of 96.48 %. We also used these corpora to study the similarity between the 5 languages at the lexical level, which is the first data-based study of the extent of 'closeness' of these languages.

## 1. Introduction

Indo-Aryan is the largest and also one of the well-studied language families in the Indian subcontinent. At the same time, it also presents one of the most controversial classification and grouping of languages, especially in terms of languages and their varieties. This is effected by two different reasons. One is the difficulty in tracing the historical path of the Indo-Aryan languages (see Masica, 1993 for detailed discussion on the problem areas and Grierson, 1931; Chatterjee, 1926; Turner, 1966; Katre, 1968; Cardona, 1974 and Mitra and Nigam, 1971 for their somewhat incompatible classification of Indo-Aryan genealogy). The second and more immediate reason is the imposition of Modern Standard Hindi (MSH) over what is now popularly known as 'Hindi Belt' and what has historically been established as a rather complex dialect continuum, with several languages and varieties being spoken in different domains of usage (see Gumperz, 1957, 1958 for an exposition of the different levels of language spoken in the area; King, 1994; Khubchandani, 1991, 1997 for a discussion on the process and resultant of this imposition; also Deo, 2018 for a brief but excellent discussion of issues surrounding the Indo-Aryan languages). Masica (1993) had established the boundaries of this continuum as starting from the language group Rajasthani on the Western side (spoken in the Western state of Rajasthan) to the language group Bihari on the Eastern side, covering other languages like Awadhi, Braj and Bhojpuri. However, with the introduction of MSH as the standard, the situation has become rather complex in the region. As Deo (2018) puts it,

*"This top-down state-imposed linguistic norm of Modern Standard Hindi (MSH) has had far-reaching effects on the dialectal situation in the Hindi belt. Until constitutional sanction for Hindi, there were relatively few native speakers of this language in either its deliberately crafted literary version, or its dialectal base, Khari Boli."*

However, with MSH being propagated and imposed as the standard through education, media, etc, there has been an increase in both the 'monolingual' speakers of MSH (largely urban, educated population, who no longer speak their parents' language) and 'bilingual' (or even multilingual) speakers who speak one of the languages of the region (with / without an understanding that it is a 'non-standard' variety of MSH) as well as MSH in different domains of usage (Khubchandani, 1997). In addition to this, it must also be noted that the speakers of MSH do not actually speak the same variety of MSH; rather they generally speak some kind of a 'mixed' variety which borrows heavily from their regional language (notwithstanding whether they speak that language or not) and it is these which could actually be called 'varieties' of MSH (see Kumar, Lahiri and Alok, 2013, forthcoming for discussion of one such variety of MSH).

Given this, there are 2 major motivations for working on the languages of the 'Hindi Belt'. The first motivation is technological. Since most of the speakers in the belt are bi-/multi-lingual (in MSH or a variety of it and at least one other language), the actual language usage is marked by code-mixing and code-switching among these languages / varieties. Now even for the most basic task like building a corpus from social media requires that these languages be automatically recognised since there might be a few users who would be writing in Bhojpuri or Magahi and others who might be writing in MSH (or one of its varieties). The second motivation is more theoretical / linguistic. We would like to explore the hypothesis about 'dialect continuum' as well as look at 'similarity' / 'closeness' of the different 'discrete' languages / varieties in this continuum using a data-based approach and give empirical evidence for or against the hypothesis that these languages form part of a 'dialect continuum'.

In this paper, we take into consideration 5 languages of the continuum – Braj, Awadhi, Bhojpuri, Magahi and MSH. Braj is spoken in Western Uttar Pradesh, Awadhi is spoken in Eastern / Central Uttar Pradesh, Bhojpuri in Eastern Uttar Pradesh and Western Bihar and Magahi is spoken in South / Central Bihar. MSH or one of its varieties is now spoken across the belt but it has its dialectal base in Khari Boli which is spoken in Western Uttar Pradesh. Thus in this group, MSH, assuming it to be a variety of Khari Boli, is the Westernmost language, followed by Braj and Magahi is the Easternmost language, with these languages also forming a continuum (shown in Fig 1 below). We will discuss the development of corpus for each of these languages and also give a basic analysis of lexical similarity among these languages. We also discuss the development of a baseline automatic language identification system for these 5 languages using the above-mentioned corpus.

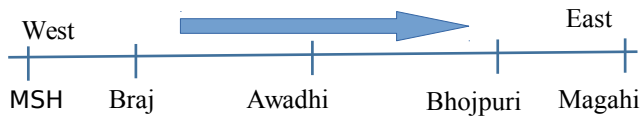

Fig 1 : Position of the 5 languages in the continuum

## 2. Related Work

Language Identification was generally considered a solved problem with several classifiers for discriminating between languages performing almost perfectly, when trained with word and character-level features. However, in the past few years, inability to replicate similar results in discriminating between varieties and closely-related languages has opened up new questions for the field and kickstarted fresh attempts at solving this problem.

Ranaivo-Malançon (2006) presents one of the first studies that tries to discriminate between two similar languages – Indonesian and Malay – using a semi-supervised model, trained with frequency and rank of character trigrams, lists of exclusive words and the format of numbers (which differed in the two languages in the sense that Malay used decimal points whereas Indonesian uses commas).

Ljubešić et al. (2007) worked on the identification of Croatian texts in comparison to Slovene and Serbian, and reports a high precision and recall of 99%.They made use of a 'black' list of words that increases the performance of the system significantly. This method was further improved by Tiedemann and Ljubešić (2012) improved this method and applied to Bosnian, Croatian and Serbian texts.

Zampieri and Gebre (2012) identified two varieties of Portuguese (Brazilian and European). They used journalistic texts for their experiments and their system gave an impressive 99.5% accuracy with character n-grams. A similar method was later used for classifying Spanish texts that used part-of-speech features, along with character and word n-grams for classification (Zampieri et al., 2013).

Xu, Wang and Li (2016) discusses the development of a system for 6 different varieties of Mandarin Chinese spoken in Greater Chinese Region. They trained a linear SVM using character and word n-grams and also word alignment features. The best system gave an accuracy of 82% which could be explained by the fact that some of the dataset that they used was noisy and also the fact that these are varieties of the same language and expected to be very close to each other, resulting in a difficulty in discriminating among them.

More recently, there have also been an increase in studies focussing on language identification on social media, specially Twitter (Williams and Dagli, 2017; Castro, Souza and de Oliveira, 2016; Radford and Gallé, 2016; Ljubešić and Kranjčić, 2015).

Ljubešić and Kranjčić (2015) worked on 'user-level' language identification instead of 'tweet-level' in which they reached an accuracy of ~98% using a simple bag-of-words model with word unigram and 6-grams and character 3-grams and 6-grams, while classifying 4 very similar South-Slavic languages – Bosnian, Croatian, Montenegrin and Serbian.

Radford and Gallé (2016) makes use of both the language as well as graph properties of tweets for discriminating between 6 languages of tweets – Spanish, Portuguese, Catalan, English, Galician and Basque – and achieved a best score of 76.63%.

Castro et al. (2016) tries to discriminate between Brazilian and European Portuguese on Twitter. They use an ensemble method with character 6-grams and word uni and bigrams to achieve a score of 0.9271.

Williams and Dagli (2017) uses geo-location, Twitter LID labels and editing by crowdsourcing to quickly annotate tweets with their location and then train a classifier using MIRA algorithm to discriminate Indonesian and Malay in tweets. They achieved an accuracy of 90.5% when trained on 1,600 tweets. Their experiments show the utility of using geo-bounding of tweets based on the location of their posting.

Despite this heightened interest in discriminating between similar languages in the European, Asian and also Arabic context, there is hardly any similar attempts to identify Indian languages. Murthy and Kumar (2006) is the only work that we came across for Indian languages. They developed pairwise language identification system for 9 Indian languages from 2 different language families – Hindi, Bengali, Marathi, Punjabi, Oriya (all from Indo-Aryan language family), Telugu, Tamil, Malayalam and Kannada (all from Dravidian language family). Given the fact that these languages are quite distinct from each other and it was a binary classification task (for each pair of languages), the classifier performed almost perfectly which was at par with most of the other state-of-the-art systems available.

Indhuja et al. (2014) also discusses identification of 5 Devanagari-based languages – Hindi, Sanskrit, Marathi, Nepali and Bhojpuri – using character and word n-grams. Even though they claim that these are similar languages, it is not the case despite the fact that they belong to the same language family and use the same script. The system gives a best performance of 88% which is not at par with the performance of modern language ID systems, mainly because of the approach that they took for solving the problem.

Aside from these, there have been hardly any attempt at automatically identifying languages in a multilingual document in Indian languages. Given the fact that most of the documents produced today are multilingual (and they do not include only one of these major languages), with social media enhancing this challenge by several fold, even for the basic task of automated data collection for Indian languages, it is imperative that automatic language identification systems be developed for Indian languages, especially closely-related languages and varieties. In this paper, we present the first attempt towards automatic identification of 5 closely-related Indian languages.

## 3. Corpus Collection

As we have already discussed above, the 4 languages that we would be working with, share a very complex and, lot of times, hierarchical relationship with MSH. As such, while we have comparatively huge amount of data available for MSH, there is hardly anything available for the other 4 languages – Braj, Bhojpuri, Awadhi and Magahi – in the written form and virtually nothing in the digitised form. The data collection process for all the 4 languages largely followed a similar methodology.

Even though the four languages mentioned above are not used in education or for official purposes, they have a very rich literary tradition[1]. In order to preserve, promote, publish and popularise literary tradition of these languages, local state governments have set up special bodies for some of these languages. For Magahi and Bhojpuri, there are Magahi Akademi (Magahi Academy) and Bhojpuri Akademi (Bhojpuri Academy) in Patna (the capital of the state of Bihar) and for Braj, there is Braj Bhasha Akademi (Braj Bhasha Academy) in Jaipur (the capital of the state of Rajasthan). Uttar Pradesh Hindi Sansthan (Uttar Pradesh Hindi Institute) performs a similar role for Awadhi. Along with these, some individuals, local literary and cultural groups and language enthusiasts also bring out publications in these languages.

Our data collection process mainly consisted of looking for printed stories, novels and essays either in books, magazines or newspapers, scanning those, running an OCR and finally proofreading the OCRed texts by the native speakers of the respective languages.

Since there is no specific OCR available for these languages, we made use of Google's OCR for Hindi that they provide in the Drive API. Since all the languages used Devanagari, we expected the OCR to give a reasonable accuracy and, barring a few times (which was due to the choice of font instead of the language *per se*), it worked rather well. This method helped us in quickly creating the corpus without the need of typing out the whole text.

In addition to this, we also managed to get some blogs in Magahi and Bhojpuri. We crawled these blogs using an in-house crawler built using Google's API. The data for MSH was also crawled from blogs using the same crawler. More details about the sources of data is given in the following subsections

### 3.1 Awadhi

As mentioned above, the data for Awadhi has been collected from Uttar Pradesh Hindi Sansthan's Library and other publication houses in Lucknow. At present, we managed to get 3 novels written in modern Awadhi -

- Chandawati
- Nadiya Jari Koyla Bhai
- Tulsi Nirkhen Raghuvar Dhama

The current corpus contains data from these 3 novels.

### 3.2 Bhojpuri

In its current form, the data for Bhojpuri is crawled from 4 different blogs -

- Anjoria
- TatkaKhabar
- Bhojpuri Manthan
- Bhojpuri Sahitya Sarita

We have also got several short story collections, novels and other literary works from Bhojpuri Akademi at Patna and we are in the process of adding those to the present corpus.

### 3.3 Braj

The data for Braj was collected from 2 main sources – Braj Bhasha Akademi in Jaipur and Braj Shodh Sansthan Library in Mathura. We got two kinds of printed literature from the two sources. We got the following from the Akademi -

- Modern fictional literature published as novel and short stories.
- Critical essays on literature (including a 12-volume set of books on the history of Braj literature, written in Braj)
- Several volumes of a magazine called 'Brajshatdal' consisting of memoirs, short stories, essays and articles.

We got the following from the library in Mathura

- Religious commentaries and essays published as books.

Our corpus currently contains data from each of these sources in almost equal proportion.

### 3.4 Magahi

Magahi data is mainly collected from 3 sources (see Kumar, Lahiri and Alok 2012, 2014 for more details) -

- Magahi Akademi at Patna : We got several novels, plays and short stories from the Akademi
- Local fieldwork in the areas of Gaya and Jehananbad : We got volumes of 2 Magahi magazines with essays and stories and also a collection of Magahi folktales.
- Crawling the Magahi blogs : The Magahi blogs mainly consist of original and translated literature in Magahi. It must be mentioned that one of the blogs contain a large dictionary of Magahi with example sentences. We have included these example sentences also in our corpus.

Currently the corpus consists of data from all these sources.

### 3.5 Modern Standard Hindi (MSH)

There are several corpora already available for MSH (Kumar, 2014a, 2014b, 2012 ; Chaudhary and Jha, 2014 and several others). However, in order to keep the domain same as that of other languages, we collected data from blogs that mainly contain stories and novels. Thus the MSH data collected for this study is also from the domain of literature.

---

[1] It must be mentioned here that it is this rich literature of these and other neighbouring languages that have been co-opted by MSH as 'Hindi' literature and is now forms part of what is known as the tradition of Hindi literature.

The present statistics for each language is summarised in Table 1

| Language | Sentences (approx.) |
|---|---|
| Awadhi | 15,000 |
| Braj | 30,000 |
| Magahi | 170,000 |
| Bhojpuri | 62,000 |
| MSH | 30,000 |

Table 1: Corpus statistics for different languages

### 4. Lexical Overlap and Distance

As we mentioned above, there have been no prior study on the similarity of these languages. So we wanted to explore it using the data that we had. Also since we wanted to build a language identification system, an exploration into the lexical overlap and similarity of these languages would have helped us predict what might be the most useful and productive way of approaching the problem. The overlap matrix (based on lexical overlap in our corpus) is given in Table 2 below.

|  | MSH | Braj | Awadhi | Bhojpuri | Magahi |
|---|---|---|---|---|---|
| **MSH** | **31,268** | 5,721 | 4,341 | 6,441 | 4,803 |
| **Braj** | 5,721 | **23,918** | 4,466 | 5,077 | 4,195 |
| **Awadhi** | 4,341 | 4,466 | **16,977** | 4,209 | 3,622 |
| **Bhojpuri** | 6,441 | 5,077 | 4,209 | **24,254** | 5,538 |
| **Magahi** | 4,803 | 4,195 | 3,622 | 5,538 | **21,791** |

Table 2: Lexical Overlap Matrix across the 5 languages

This lexical overlap was calculated using a subset of 10,000 sentences of each language, with a total of 50,000 sentences (adding up to a variable number of unique tokens – represented in the overlap matrix above) from the corpora. As you would notice, the results are largely on the expected lines with languages closer together depicting greater overlap. Barring Bhojpuri which shares maximum overlap with MSH (even though they are not the neighbouring languages in the continuum) and least with Awadhi (which is closest to it), all other languages depict the expected behaviour. For example, Awadhi and Braj depict least overlap with Magahi as they are far apart in the continuum. Similarly, Magahi shares maximum overlap with Bhojpuri, which is closest to it. In general, we think the overlap is pretty high (upto 25% of tokens, at times) given the fact that this was a completely naive calculation that was carried out without any normalization of data. This implies that if the languages share the same root but applies different set of morphemes to those roots (which is quite often the case) then that are still considered non-overlapping. It is only when the tokens exactly match that they are considered overlapping. The same calculation with more sophisticated techniques might result in higher overlap numbers.

In addition to this word-level analysis, we also carried out a more nuanced character-level analysis using Levenhestein Edit Distance between words of the two languages. We calculated the edit distance between every pair of words in every pair of language and averaged those out to calculate an average 'distance' between the two languages. Since we do not have a standard way of calculating distance between two languages, we have taken edit distance as the proxy for that. The results are summarised in the form of a distance matrix in Table 3 below. The top row for each language in the table shows an overall edit distance while the bottom row shows the length-controlled edit distance.

As we could see, in general, the edit distance between any pair of languages is quite high and not very far apart from each other. Despite the averages being not very apart from each other, we could see a general trend of average edit distance increases slightly as they become farther in language continuum. Also like in the case of word overlap, all the languages have greatest edit distance from MSH and among those Magahi has the greatest edit distance from MSH while Braj has the least. Similarly, Awadhi and Braj has the smallest edit distance. If we control for the length of the words such that we calculate the edit distance when the length of the words are equal, the overall edit distance is approximately 1 point lower but the trends are still similar.

|  | MSH | Braj | Awadhi | Bhojpuri | Magahi |
|---|---|---|---|---|---|
| **MSH** | 0 | 6.792 | 6.823 | 6.880 | 6.987 |
|  | 0 | 5.853 | 5.375 | 5.493 | 5.549 |
| **Braj** | **6.792** | 0 | 6.249 | 6.323 | 6.433 |
|  | **5.853** | 0 | 5.302 | 5.432 | 5.488 |
| **Awadhi** | 6.823 | **6.249** | 0 | 6.347 | 6.455 |
|  | 5.375 | **5.302** | 0 | 5.447 | 5.496 |
| **Bhojpuri** | 6.880 | 6.323 | 6.347 | 0 | 6.518 |
|  | 5.493 | 5.432 | 5.447 | 0 | 5.481 |
| **Magahi** | **6.987** | 6.433 | 6.455 | 6.518 | 0 |
|  | **5.549** | 5.488 | 5.496 | 5.481 | 0 |

Table 5 : Average edit distance across the 5 languages

# 5. Language Identification: Experiments and Results

We use a total dataset of 10,000 sentences in each of MSH, Braj, Bhojpuri and Magahi and 9,744 sentences in Awadhi, taken from the corpora discussed above, for developing a sentence-level language identification systems for the 5 languages. We divide the dataset into train:test ratio of 80:20. The train set is used for training a Linear SVM classifier using 5-fold cross-validation. We tune only C hyperparamter of the classifier and arrive at the best classifier using Grid Search technique. We use scikit-learn library (in Python) for all our experiments.

Based on the results obtained in previous studies and their robustness in language identification tasks, we experimented with the most basic frequency distribution of character and word n-gram features for the problem.

**Character n-gram features**: We used character bigram (CB), trigram (CT), four-grams (CF) and five-grams (CFI) and their different combinations in our experiments

**Word n-gram features**: We used word unigrams (WU), bigrams (WB) and trigrams (WT) and their combinations for our experiments.

**Combined features**: We also experimented with different combination of both of the above features.

We used the frequency of each feature as the feature values.

| Features | Precision | Recall | F1 | Accuracy |
|---|---|---|---|---|
| **Character n-gram features** | | | | |
| CB+CT (C1) | 0.96 | 0.96 | 0.96 | 95.868 |
| CB+CT+CF (C2) | 0.96 | 0.96 | 0.96 | 96.422 |
| CB+CT+CF+CFI (C3) | 0.96 | 0.96 | 0.96 | **96.482** |
| **Word n-gram features** | | | | |
| WU (W1) | 0.79 | 0.78 | 0.79 | 78.361 |
| WU+WB (W2) | 0.8 | 0.79 | 0.79 | **79.167** |
| WU+WB+WT (W3) | 0.8 | 0.79 | 0.79 | 78.744 |
| **Combination of character and word n-gram features** | | | | |
| C1+W1 | 0.96 | 0.96 | 0.96 | 95.878 |
| C1+W2 | 0.96 | 0.96 | 0.96 | 95.848 |
| C1+W3 | 0.96 | 0.96 | 0.96 | 95.827 |
| C2+W1 | 0.96 | 0.96 | 0.96 | 96.372 |
| C2+W2 | 0.96 | 0.96 | 0.96 | 96.362 |
| C2+W3 | 0.96 | 0.96 | 0.96 | 96.331 |
| C3+W1 | 0.96 | 0.96 | 0.96 | 96.472 |
| C3+W2 | 0.96 | 0.96 | 0.96 | **96.483** |
| C3+W3 | 0.96 | 0.96 | 0.96 | 96.472 |

Table 4 : Performance of different feature sets on test set

The performance of the different models over the test set is summarised in Table 4.

As we could see, a combination of character bi-gram to 5-grams give the best result of 96.48%. However, the word n-gram features do not seem to work at all. Individually, they give a below par accuracy of 79.16%. This was anticipated and could be explained by the fact that these languages share a large amount of their vocabulary – upto 25% even when going by the strictest evaluation in terms of word-forms – and so they may not act as discriminating features for the classification task. On the other hand, we have also seen that the average edit distance between the words of two languages is close to 7, thereby, implying a greater dissimilarity at character-level, which might be helpful in classification task. Also a combination of character bigrams to 5-grams prove to be most useful, which is consistent with the previous experiments for different languages. As is evident, combining character n-grams with word n-grams does not lead to a better result, probably, because character n-grams already capture the discriminating features of word n-grams.

Table 5 below gives a summary of the language-wise performance of the classifier on the test set. It seems that Magahi is the best-performing language while MSH is the worst one with almost 3 point decrease in F1 score. Awadhi shows the worst precision and best recall which implies that further optimization on Awadhi might lead to a better result.

| | Precision | Recall | F1 | Test Samples |
|---|---|---|---|---|
| **MSH** | 0.96 | 0.95 | **0.95** | 1996 |
| **Braj** | 0.97 | 0.95 | 0.96 | 1976 |
| **Awadhi** | **0.94** | **0.98** | 0.96 | 1986 |
| **Bhojpuri** | 0.98 | 0.97 | 0.97 | 1995 |
| **Magahi** | 0.98 | 0.97 | **0.98** | 1969 |

Table 5: Language-wise performance

The confusion matrix in Table 6 shows that misclassification of MSH and Braj as Awadhi is the major source of low precision of Awadhi. Similarly, MSH is misclassified as Braj and Awadhi while Braj is also often misclassified as MSH. What is noticeable is that languages that are far apart in the continuum are *not* confused with each other. So for example, there is hardly any instance of Braj, Awadhi or even MSH being misclassified as Magahi while Bhojpuri is more often taken as Magahi by the classifier (and it holds also the other way round). Moreover, Bhojpuri, Braj and Awadhi are all classified as MSH quite significant number of times. This is despite the fact that Bhojpuri is not at all close to Hindi in the dialect continuum. This could be indicative of the greater influence of MSH over the languages across the so-called 'Hindi Belt' and a slow convergence of these languages into MSH, which has resulted in greater 'closeness' of all the languages with MSH.

|          | MSH  | Braj | Awadhi | Bhojpuri | Magahi |
|----------|------|------|--------|----------|--------|
| MSH      | 1895 | 33   | 42     | 16       | 10     |
| Braj     | 35   | 1887 | 46     | 2        | 6      |
| Awadhi   | 19   | 20   | 1943   | 2        | 2      |
| Bhojpuri | 23   | 8    | 11     | 1930     | 23     |
| Magahi   | 5    | 5    | 16     | 25       | 1918   |

Table 6 : Confusion Matrix on the test set

A closer look at the errors made by the system reveal that quite a few of these errors are because of the noise in the test set. Some are errors because of named entities – strictly speaking this cannot be classified as an error since names are shared across the languages. However, there are some errors which are because of the sufficient overlap between the two languages or availability of sufficient discriminating features in the test sample as in the following example -

1. उ कतल करे जानें, तब्बो फूल बरसावे ले जनता

Even when he keep on killing, the public will still adore him. [Predicted: **Magahi**; Actual: **Bhojpuri**]

2. पत्रिका में अच्छा जीवन चरित सुरेश दुबे 'सरस' जी के लिखे के बात तय होल।

The good life skecth in the magazine was decided after the writing of Suresh Dubey 'Saras' ji [Predicted: **Bhojpuri**; Actual: **Magahi**]

3. अभी बहुत काम है ।

Now there is lot of work left to be done [Predicted: **Awadhi**; Actual: **Hindi**]

4. खैर युद्ध के अवशेष पिंकी देवी बाहर फेंकि आईं ।

Anyway, Pinkey Devi threw out the remains of the war. [Predicted: **Hindi**; Actual: **Awadhi**]

Most of the time, the verbal endings provide a strong clue towards the actual language of the sentence. However, in the examples, given above either the verb is missing (e.g. 1) or it is shared with the other languages (e.g. 2 and 3). In example 3, the sentence is quite short one and so there is not enough discriminating feature available for the classifier. Example 4 can be explained by he use of borrowed words like 'युद्ध' and 'अवशेष', coupled with a verb which is common in Hindi led to it being classified as Hindi.

A lot of these errors could be handled with the use of language-specific morphological features as well as by removing the noise in the data (which, in any case, is not large in number). However, it will remain a difficult task to classify the language in the case of really small sentences or a large amount of overlapping lexicon.

## 6. Summing Up

In this paper, we have discussed the creation of a comparable corpus of 5 closely-related Indian languages – MSH, Braj, Awadhi, Bhojpuri and Magahi. It need not be mentioned that not only language resources and technologies but even published material (digital / online or in print) in these languages are very scarce and it is a very resource-intensive process to create corpora for these languages. Moreover it is also not possible to crawl data from the web or social media because often these languages are mixed together and it would require manual intervention to segregate those. Based on the corpus developed, we have presented a basic analysis of lexical overlap and average edit distance of these languages, which might be useful from a variationist / sociolinguistic point of view. The analysis shows that there is indeed a greater distance in between languages that are geographically apart, thereby, providing an empirical evidence of a dialect continuum. At the same time, it also provides a strong argument against positing these languages as a variety of one language – they are sufficiently different from each other to be posited as distinct, discrete points in the continuum. We have also developed a baseline automatic language identification system, which is the first such attempt for closely-related Indian languages. The system currently gives an accuracy of over 96% with character 5-grams and may be taken as a baseline for future experiments towards automatically discriminating among these languages.

## 7. Bibliographical References